# A Resampling-Based Framework for Network Structure Learning in High-Dimensional Data

Ziwei Huang[1,5,*], Zeyuan Song[2,3], Paola Sebastiani[2,3,4], Stefano Monti[5,6,7,*]

[1] Department of Physics, Boston University, Boston, MA,
[2] Institute for Clinical Research and Health Policy Studies, Tufts Medical Center, Boston, MA
[3] Department of Medicine, School of Medicine, Tufts University, Boston, MA
[4] Data Intensive Study Center, Tufts University, Boston, MA
[5] Division of Computational Biomedicine, Boston University Chobanian & Avedisian School of Medicine, Boston, MA
[6] Department of Biostatistics, Boston University School of Public Health, Boston, MA
[7] Bioinformatics Program, Faculty of Computing and Data Science, Boston University, Boston, MA
* {ziwhuang,smonti}@bu.edu

## Summary

*RSNet* is an open-source R package that provides a resampling-based framework for robust and interpretable network inference, designed to address the limited-sample-size challenges common in high-dimensional (e.g., 'omics') data. It supports both the estimation of partial correlation networks modeled as Gaussian networks[1] and conditional Gaussian Bayesian networks for mixed data types that combine continuous and discrete variables[2]. The framework incorporates multiple resampling strategies, including bootstrap, subsampling, and cluster-based approaches, to accommodate both independent and correlated (e.g., family-based) observations. To enhance interpretability, *RSNet* integrates graphlet-based topology analysis that captures higher-order connectivity and edge sign information, enabling single-node and subnetwork-level insights. Notably, *RSNet* is the first R package to efficiently construct signed graphlet degree vector matrices (GDVMs) in near-constant time for sparse networks, providing scalable analysis of higher-order network structure. Collectively, *RSNet* offers a versatile tool for statistically reliable and interpretable network inference in high-dimensional data.

## Statement of need

Network inference methods are widely used to model dependencies among variables in high-dimensional data, supporting discovery and hypothesis generation in diverse research domains[3–6]. Commonly applied approaches such as correlation or co-expression networks[7–9] are straightforward to implement but cannot distinguish direct from indirect dependencies[3]. In contrast, Gaussian networks, also known as partial correlation neworks,[3–5] and conditional Gaussian Bayesian networks (CGBNs)[10,11] estimate conditional dependencies, offering a higher-resolution representation of complex systems. However, the reliability of inferred network structures is often compromised by the limited sample sizes in high-dimensional data, a challenge commonly referred to as the "small $n$, large $p$" problem, where $n$ denotes the number of samples and $p$ the number of variables[3,5,6,12].

*RSNet* addresses this limitation by introducing a resampling-based framework that quantifies edge-level uncertainty and integrates information across multiple inferred networks to construct a robust consensus network. The framework supports both Gaussian networks for continuous data and CGBNs for mixed data types and can accommodate correlated or family-based observations[13]. This design provides empirical confidence intervals, adjusted p-values, and edge-selection frequencies, offering a fine-grained assessment of network reliability and structure[6,14,15].

In addition to improving reliability, *RSNet* enhances interpretability through graphlet-based topology analysis, which captures higher-order local connectivity patterns and incorporates edge sign information[16–18]. These functionalities enable detailed examination of node-level structural roles and facilitate comparative analyses between networks inferred under different conditions[19–21]. Existing R packages for network inference do not support the construction of signed graphlet degree vector matrices (GDVMs), where brute-force enumeration has complexity greater than $O(p^3)$, with $p$ denoting the network size. *RSNet* overcomes this barrier by combining state-of-the-art graphlet counting algorithms[22–24] with parallelization, establishing an efficient method to construct signed GDVMs in $O(\mid d \mid)$, where $\mid d \mid$ is the average degree, resulting in near-constant time complexity for sparse networks.

**State of the field**

*RSNet* provides a unified, resampling-based framework for network inference and analysis that supports both independent and correlated datasets. While packages with similar functionalities exist, they typically either focus on specific components of the workflow or do not support resampling-based strategies. These limitations can compromise the reliability of inferred network structures in high-dimensional settings with limited sample sizes and noisy data, as well as the interpretability of downstream analyses. In addition, the lack of support for cluster-based resampling approaches can lead to underestimation of variability, hence a false positive inflation, when analyzing correlated or family-based data[13].

For Gaussian network structure inference, packages such as *glasso*[5] and *huge*[25] provide efficient estimation of precision matrices and corresponding network structures, but do not offer statistical inference (e.g., adjusted p-values and/or confidence intervals) for individual edges. Methods implemented in *SILGGM*[14] extend this framework by providing edge-level inference through asymptotic normality approximations, enabling the estimation of adjusted p-values and confidence intervals. However, these approaches rely on single-network estimation and do not incorporate resampling-based strategies to assess stability. *BDgraph*[26] uses Bayesian approaches to estimate network structures by sampling from the posterior distribution and to provide edge inclusion probabilities. While these quantify uncertainty in a Bayesian sense, they are not directly comparable to frequentist measures and primarily emphasize network structure rather than edge-level statistical significance.

For conditional Gaussian Bayesian networks, *RHugin*[12]implements efficient structure learning (e.g., PC algorithm) for mixed data types. However, it does not support resampling-based strategies and does not provide measures of stability such as edge selection frequencies or frequencies of higher-order dependencies (e.g., Markov blankets[27]).

For downstream network analysis, *igraph*[28] provides a comprehensive suite of tools, including centrality analysis and community detection. However, it does not support graphlet-based analysis or the construction of GDVMs. *ORCA*[24] provides one of the most efficient implementations for GDVM construction but is limited to unsigned networks and does not support signed graphlet analysis.

*RSNet* addresses these limitations by integrating existing methods within a resampling and parallelized framework, providing edge-level empirical confidence intervals, adjusted p-values, and edge-selection frequencies. This approach offers a fine-grained assessment of network reliability while maintaining computational efficiency. In addition, *RSNet* offers a suite of tools for downstream analysis, unifying network inference and structural interpretation within a single workflow implemented in an open-source R package.

**Software design**

*RSNet* is designed as a modular, resampling-based framework that enables flexible, end-to-end workflows from network inference to higher-order structural analysis. For Gaussian networks, these estimates include empirical confidence intervals and nominal or adjusted p-values; for conditional Gaussian Bayesian networks, they correspond to edge-selection frequencies. The resulting consensus network forms the basis for downstream analyses such as centrality analysis, community detection, graphlet-based topology analysis, and differential connectivity analysis (Figure 1).

A central design choice is the use of resampling-based strategies instead of single-network estimation. This approach explicitly quantifies edge-level uncertainty and improves robustness in high-dimensional, limited-sample-size settings. *RSNet* implements multiple resampling strategies. For both Gaussian networks and conditional Gaussian Bayesian networks (CGBNs), users can choose among four general approaches: (1) unstratified bootstrap, (2) unstratified subsampling, (3) stratified bootstrap, and (4) stratified subsampling, depending on data characteristics and study design. In addition, *RSNet* supports cluster-based resampling methods for correlated or family-based datasets in Gaussian networks, including (1) cluster bootstrap, which samples entire clusters with replacement to preserve intra-cluster dependencies, and (2) fractional cluster bootstrap, which samples a subset of clusters with replacement. These procedures are implemented in the function "ensemble_ggm()", which leverages inference algorithms from the *SILGGM* package[14]. For CGBNs, the function "ensemble_cgbn()" provides analogous resampling-based network inference using algorithms from the *RHugin* package[12].

Parallelization is integrated as a core design principle across both network inference and higher-order topological analysis. In addition to accelerating the resampling-based inference pipeline, parallel computing is extended to the construction of GDVMs, enabling scalable analysis of graphlet-based representations that are otherwise computationally intensive. Beyond network inference, *RSNet* supports a range of downstream analyses on the consensus network, including graphlet-based topology analysis, differential connectivity analysis, centrality analysis, and community detection, providing a unified framework for both statistical inference and structural interpretation.

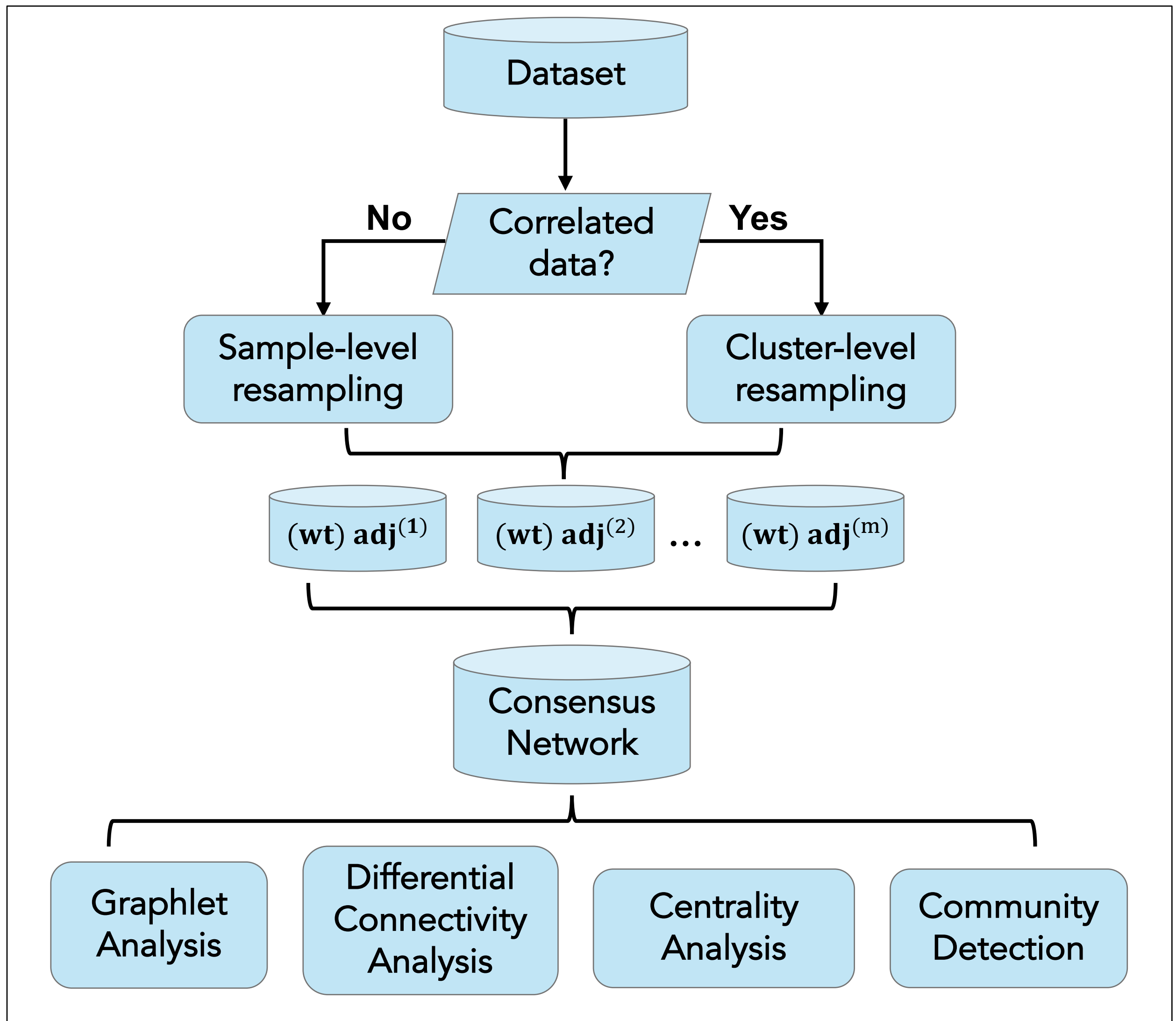

**Figure 1: Overview of RSNet.** RSNet accepts a sample-by-feature dataset and allows users to specify whether observations are independent or correlated. The resampling framework generates an ensemble of weighted or binary adjacency matrices over m iterations, which are subsequently integrated into a consensus network that serves as the basis for downstream analyses.

## Research impact statement

*RSNet* has been applied to real-world biological datasets, including aging and longevity cohorts such as the New England Centenarian Study[29,30], the Long Life Family Study[31], and Integrative Longevity Omics[31]. In addition, *RSNet* has been applied to disease-related datasets, including late-onset Alzheimer's disease (LOAD)[32] and cancer cohorts from The Cancer Genome Atlas (TCGA)[33]. These applications demonstrate the utility of the framework for robust network inference and downstream topological analysis across diverse biological settings. *RSNet* is released as open-source software with documented functions and reproducible workflows to support transparent and reliable use.

## Discussion

*RSNet* provides a versatile and scalable R package for resampling-based network inference, designed to address the challenges of limited sample size in high dimensional datasets. By integrating both Gaussian networks and conditional Gaussian Bayesian networks, *RSNet* supports structure learning for continuous and mixed data types within a unified framework.

The package enhances interpretability by integrating standard network analysis tools with graphlet-based methods for higher-order topological characterization. To the best of our knowledge, *RSNet* is the first R package to implement the construction of a signed GDVM in approximately constant time for sparse networks.

To conclude, *RSNet* facilitates reproducible, statistically robust, and interpretable network analysis. Its modular and parallelized design supports large-scale applications while maintaining transparency and flexibility, making it a user-friendly open-source resource for high-dimensional network inference and comparative structural analysis.

## AI usage disclosure

Generative AI tools were used to assist with minor aspects of manuscript preparation, including language editing and code formatting. All scientific content, methodological design, and implementation were developed by the author. Any AI-assisted outputs were carefully reviewed and validated to ensure accuracy and consistency with the intended methods and results.

## Availability

The latest version of the *RSNet* package along with additional information on the installation process can be found on github.com/montilab/RSNet (DOI: 10.5281/zenodo.20122935).

## Funding

This work was supported in part by the National Institutes of Health, NIA cooperative agreements U19 AG023122-16 and UH3 AG064704, and NIDCR R01 R01DE031831. The findings and conclusions presented in this paper are those of the author(s) and do not necessarily reflect the views of the NIH.